\documentclass[conference]{IEEEtran}
\IEEEoverridecommandlockouts
\usepackage{cite}
\usepackage{amsmath,amssymb,amsfonts}
\usepackage{algorithmic}
\usepackage{graphicx}
\usepackage{textcomp}
\usepackage{xcolor}
\def\BibTeX{{\rm B\kern-.05em{\sc i\kern-.025em b}\kern-.08em
    T\kern-.1667em\lower.7ex\hbox{E}\kern-.125emX}}
\begin{document}

\title{Traffic Sign Detection With Event Cameras and DCNN}
\author{
\IEEEauthorblockN{Piotr Wzorek}
\IEEEauthorblockA{\textit{Embedded Vision Systems Group}, \\ \textit{Computer Vision Laboratory}, \\ \textit{Department of Automatic Control and Robotics}, \\ \textit{AGH University of Science and Technology}, \\
\textit{Kraków, Poland} \\
\textit{pwzorek@agh.edu.pl}}
\and
\IEEEauthorblockN{Tomasz Kryjak, Senior Member, IEEE}
\IEEEauthorblockA{\textit{Department of Digital Systems} \\ \textit{Silesian University of Technology}, \\
\textit{Gliwice, Poland} \\
\textit{tomasz.kryjak@agh.edu.pl}}
}
\maketitle

\begin{abstract}
In recent years, event cameras (DVS -- Dynamic Vision Sensors) have been used in vision systems as an alternative or supplement to traditional cameras. 
They are characterised by high dynamic range, high temporal resolution, low latency, and reliable performance in limited lighting conditions -- parameters that are particularly important in the context of advanced driver assistance systems (ADAS) and self-driving cars. 
In this work, we test whether these rather novel sensors can be applied to the popular task of traffic sign detection. 
To this end, we analyse different representations of the event data: event frame, event frequency, and the exponentially decaying time surface, and apply video frame reconstruction using a deep neural network called FireNet. 
We use the deep convolutional neural network YOLOv4 as a detector.
For particular representations, we obtain a detection accuracy in the range of 86.9-88.9\% mAP@0.5. 
The use of a fusion of the considered representations allows us to obtain a detector with higher accuracy of 89.9\% mAP@0.5.
In comparison, the detector for the frames reconstructed with FireNet is characterised by an accuracy of 72.67\% mAP@0.5. 
The results obtained illustrate the potential of event cameras in automotive applications, either as standalone sensors or in close cooperation with typical frame-based cameras.
\end{abstract}

\begin{IEEEkeywords}
event cameras, dynamic vision sensors, CNN, traffic sign detection, event data processing
\end{IEEEkeywords}

\section{Introduction}

Event cameras are neuromorphic sensors that asynchronously record changes in pixel brightness in the form of so-called events. 
They are increasingly used in the context of autonomous vehicles, due to their high dynamic range and high temporal resolution. 
Event cameras are suitable for very strong and low-light conditions and situations when the object moves very fast with respect to the sensor.
In this work, we use event data for traffic sign detection -- a task that is very relevant for autonomous vehicles, yet rather complex due to significant differences between elements of the same class of objects.


Implementing object detection for event data is problematic because state-of-the-art solutions, including deep convolutional neural networks (DCNN), are designed to work with ``classical'' video frames.
In this paper, we consider the use of different methods to represent event data in the form of frames: event frame, event rate, exponentially decaying time surface, a fusion of the mentioned representation and frame reconstruction with the use of the FireNet neural network \cite{b4}.
We use the YOLOv4 network as the detector. 
The results obtained allow us to evaluate the applicability of event cameras in traffic sign detection systems.



The main contributions of this paper can be summarised as follows:
\begin{itemize}
    \item To the best of our knowledge, this is the first publication which describes the implementation of traffic sign detection using event data and classical deep convolutional neural networks.
    \item We propose a methodology for filtering and verifying the annotation correctness of an event dataset based on video frame reconstruction and a simple classifier based on a~convolutional neural network. 
    \item We propose the apply of a fusion of different event data representations for the task of traffic sign detection using DCNN, which may be applicable also to other detection problems. 
\end{itemize}
The remainder of this paper is organised as follows.
In Section \ref{ch:cameras} we introduce event cameras along with their properties and data representation format. 
Section \ref{ch:representations} describes different event data representations in the form of event frames. 
Section \ref{ch:detection} presents the methods used in the literature for object detection based on event data, as well as the  considered traffic sign detection problem. 
Section \ref{ch:dataset} describes the dataset used, as well as its filtering and verification process.
The neural network training process is presented in Section \ref{ch:train}. 
The paper ends with a discussion of conclusions drawn from the conducted research and a description of plans for the further development of the project.

\section{Event cameras}
\label{ch:cameras}

An event camera (also known as Dynamic Vision Sensor -- DVS) is a neuromorphic sensor that takes its inspiration from the human eye \cite{davide}. 
Unlike classical cameras, which record the brightness (colour) level for a given pixel every specified time interval (frame per second parameter), a~DVS records brightness changes independently (asynchronously) for individual pixels.
Consequently, the data captured by the camera does not depend on the clock but the dynamics of the scene.
As a result, a stream of events is available on the output, where each is described by 4 values:
\begin{equation}
e = \{t, x, y, p\}
\end{equation}
where: $t$ is the event capture time (timestamp), $x$ and $y$ are the pixel coordinates, and $p$ is the value -1 or 1 corresponding to the event polarity (positive or negative change in pixel value reaching a fixed threshold). 

Such an approach has a number of interesting consequences.
First, a single pixel consists of an analogue and a digital part.
Changes are detected in the analogue part, which results in low latency (fast response time) and energy efficiency (insufficient change means that the digital part will not be triggered).
In addition, the analysis of the logarithm of the change in brightness of each particular pixel independently allows one to obtain a high dynamic range (above 120 dB). 
Contrast sensitivity of event cameras differs for particular sensors. 
Typical DVS's set thresholds between 10-50\% illumination change as too low setting can result in high number of noise events. 
Consequently, DVS sensors are relatively resistant to large differences in the brightness of the recorded scene, which for classical cameras becomes a problem (part of the image is overexposed or too dark). 
This property also enables the correct detection of objects even under very high or limited light conditions. 
Second, the transmission of information only about the occurring changes eliminates the usually unwanted redundancy (a traditional camera sends information about the status of all pixels, even those whose brightness has not changed). 
Another characteristic of event cameras is their high sampling rate -- the timestamp has a resolution of microseconds.
In addition, the camera data stream can reach up to 1200 MEPS (million events per second) depending on the chip and hardware interface used. 
This speed was reported in \cite{camera} for image resolution of $1280\times960$ with 4.95 $\mu$m pixel pitch (distance between pixels).
The limitation of redundant information and the high frequency enable the processing of key information even under conditions of high variability, i.e. when the objects registered by the camera move in relation to it at high speed. 
It is worth noting that the above is true in the so-called usual scenario. 
In the case when, due to rapid changes of brightness or dynamic movement of the camera, all pixels would start to generate events, the output data stream will be larger than for traditional cameras (for example, due to the larger representation of a single pixel).

However, it should be noted that the event camera data stream, which can be described as a sparse point cloud in space-time, presents significant challenges in terms of designing analysis and recognition algorithms.
First, the previous achievements of over 60 years of computer vision cannot be easily applied.
Three main approaches are used: an attempt to analyse the data as a point cloud (in a way similar to, e.g. LiDAR data), transformation (accumulation) of event data into various types of event frames, and reconstruction.
The latter two approaches will be discussed in more detail later in this paper.

The features of event cameras described above are crucial in the context of vision systems for autonomous vehicles or advanced driver assistance systems.
They perform well for essential vision tasks for fast-moving objects, regardless of lighting conditions and vehicle speed. 
However, event representation presents a considerable challenge for designers of vision algorithms, including object analysis and recognition.

\section{Event-frame representation}
\label{ch:representations}

Due to the non-traditional format of the data recorded by event cameras, it is difficult to use them with state-of-the-art computer vision algorithms. 
Therefore, various event data representations in the form of frame-matrices analogous to traditional video data are proposed in the literature. 

To define some of the representations, we use a notation analogous to the one used in the work \cite{b1}. 
We define a function $\Sigma_{e}$ that, for each pair of event coordinates $u(x,y)$, matches the time of its event $t$ (Equation \eqref{eq:sigma}) and a function $P_{e}$ that, for an event coordinate, matches its polarity (Equation \eqref{eq:pe}). 
\begin{equation}
\label{eq:sigma}
u: t = \Sigma_{e}(u), \Sigma_{e}: R^{2} \rightarrow R
\end{equation}
\begin{equation}
\label{eq:pe}
u: p = P_{e}(u), P_{e}: R^{2} \rightarrow \{-1, 1\}
\end{equation}
Additionally, $\tau$ was defined as the accumulation time of the event data (the representation is created through aggregation). 
During our ongoing research, the value of $\tau$ was set to 10 ms.

\subsection{Event frame}

The simplest form of event data representation is the ``event frame'' proposed in the paper \cite{b2}.
Each pixel is assigned a polarity value of the recorded events. 
Therefore, it takes into account only information about coordinates and polarisation of pixel brightness changes, ignoring temporal information. 
A representation defined in this way can be described as:
\begin{equation}
\label{eq:event_frame}
f(u,t) = \left\{ \begin{array}{lr} P_{e}(u) & for \ t-\Sigma_{e}(u)\in(0;\tau) \\ 0 & for \ t-\Sigma_{e}(u)\in(\tau;+\infty) \end{array}\right.
\end{equation}

\begin{figure}[!t]
    \centering
    \includegraphics[width=0.45\textwidth]{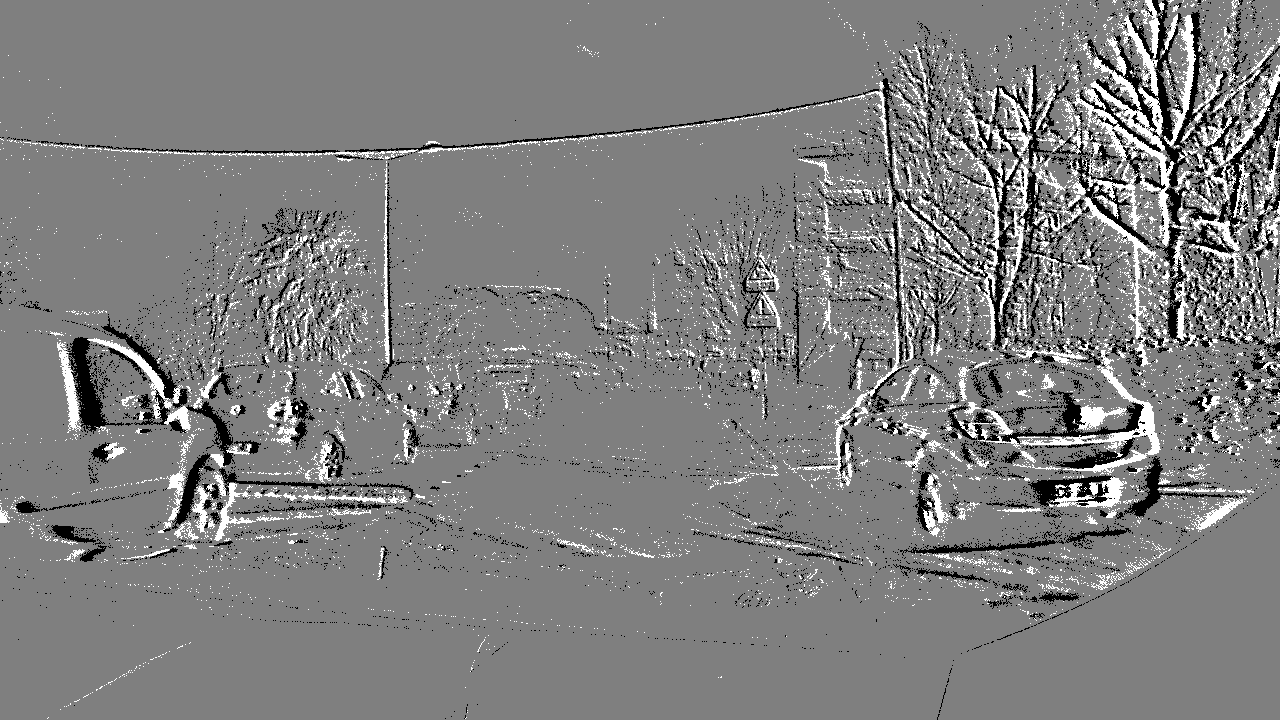}
    \caption{Event frame representation. White/black -- positive/negative event, gray -- no event.} 
    \label{fig:frame}
\end{figure}

An example event frame representation is shown in Figure \ref{fig:frame}. 
Any pixels for which no event have been registered are marked with the value 127, and events are marked as 0 or 255 depending on the polarity. 

\subsection{Exponentially decaying time surface}

\begin{figure}[!t]
    \centering
    \includegraphics[width=0.45\textwidth]{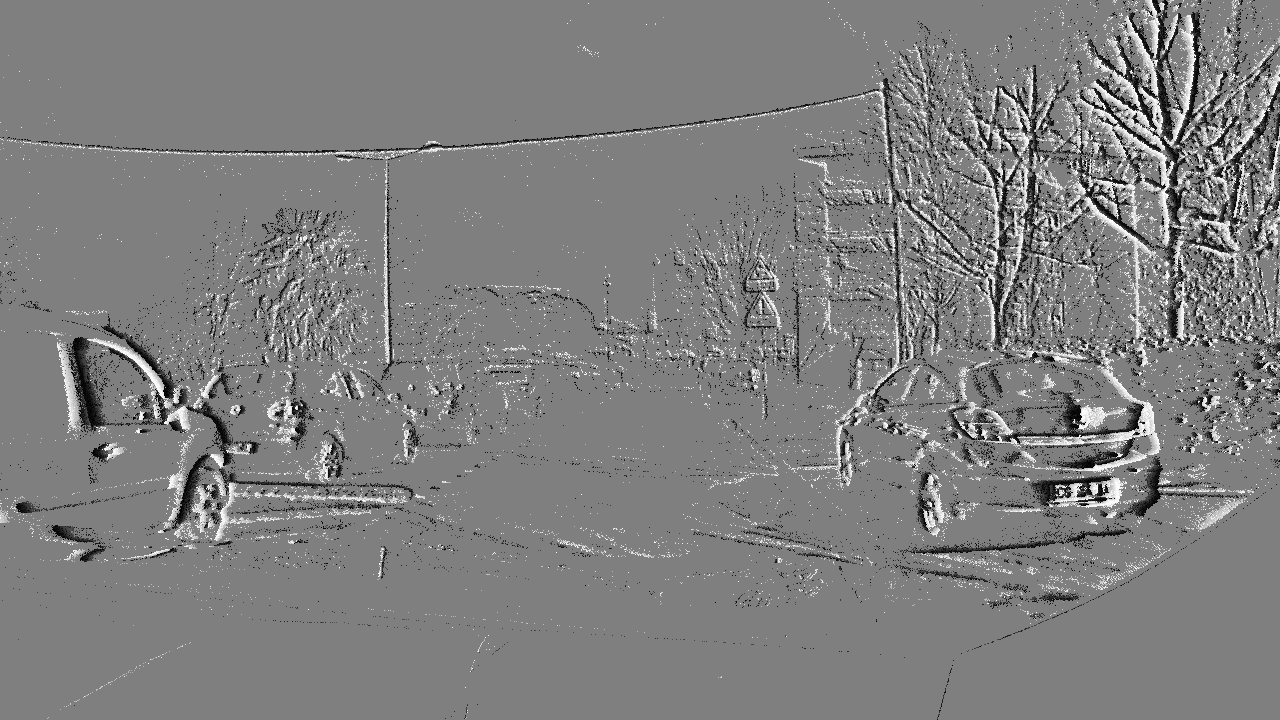}
    \caption{Exponentially decaying time surface representation} 
    \label{fig:decay}
\end{figure}

An extension of the idea of an event frame is a representation that also takes into account the time of occurrence of an event in a time window \cite{b2}. 
This method assumes that the weight of information decreases exponentially with time to zero.
This representation is denoted by the Equation \eqref{eq:decay} and is visualised in Figure \ref{fig:decay}.
\begin{equation}
\label{eq:decay}
f(u,t) = \left\{ \begin{array}{lr} P_{e}(u)*e^{\frac{\Sigma_{e}(u)-t}{\tau}} & for \ \Sigma_{e}(u) \leq t \\ 0 & for \ \Sigma_{e}(u) > t \end{array}\right.
\end{equation}

\subsection{Event frequency}
\begin{figure}[!t]
    \centering
    \includegraphics[width=0.45\textwidth]{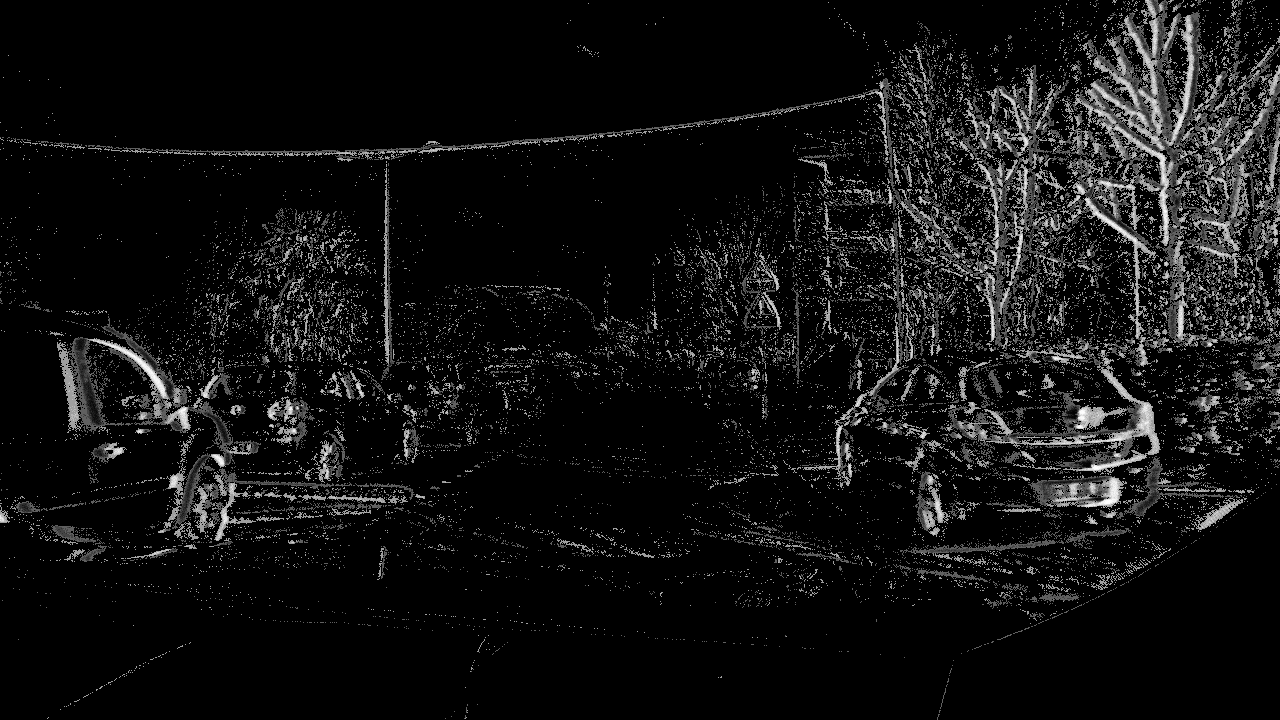}
    \caption{Event frequency representation} 
    \label{fig:sigmoid}
\end{figure}

The sigmoid representation uses information about the frequency of events for a given pixel, which is ignored by the approaches presented so far \cite{b3}. 
It is defined as:
\begin{equation}
\label{eq:sigmoid}
f(x) = 255 * \frac{1}{1+e^{-x/2}}
\end{equation}
The sum of the polarisation values present in a given pixel is denoted as $x$. 
A visualisation of this reconstruction method is presented in Figure \ref{fig:sigmoid}.

\subsection{Frame reconstruction with DNN}

\begin{figure}[!t]
    \centering
    \includegraphics[width=0.45\textwidth]{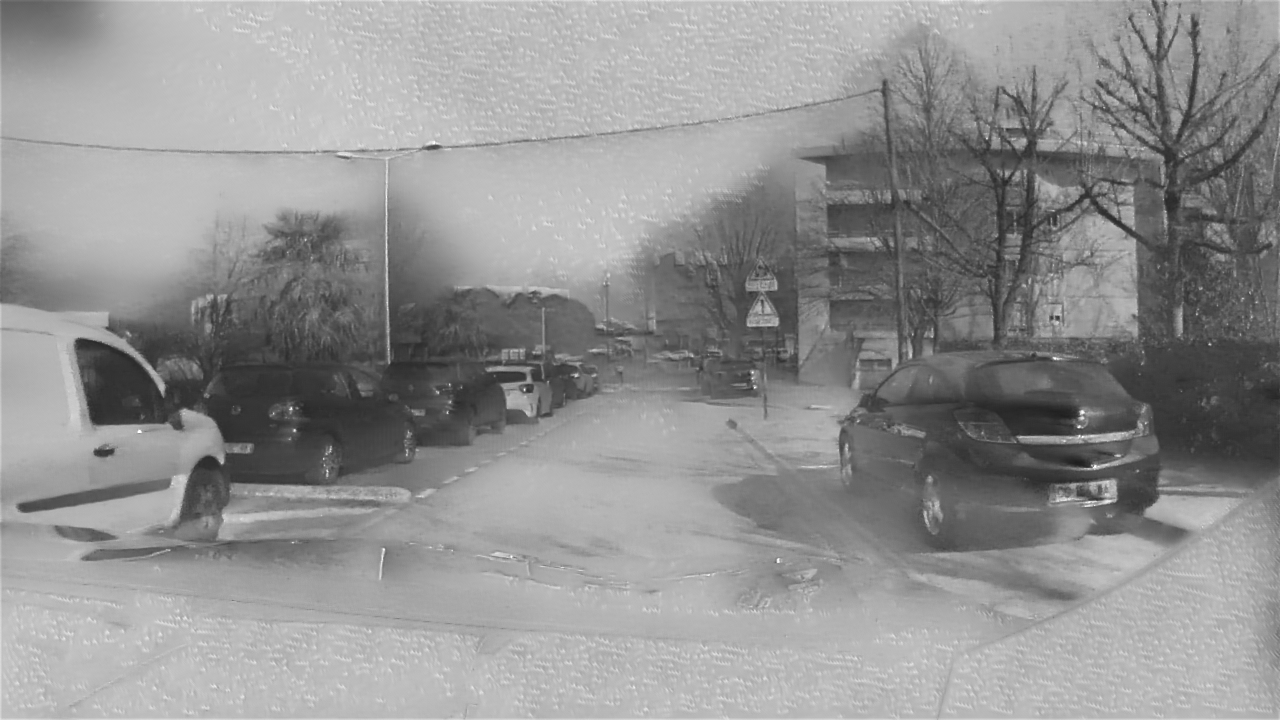}
    \caption{Frame reconstructed with FireNet} 
    \label{fig:reconstruction}
\end{figure}

In addition to aggregating events over a specific time window, another possibility is to use deep neural networks to reconstruct video frames from event data. 
An example is FireNet -- a state-of-the-art fully convolutional recurrent neural network \cite{b4}.
It is characterised by a lower computational complexity and higher performance than alternatives presented in the literature (such as E2VID \cite{b5}). 
A sample reconstruction is presented in Figure \ref{fig:reconstruction}.


\section{Object detection for event data}
\label{ch:detection}
\subsection{Previous work}

Event cameras, due to the properties discussed in Section \ref{ch:cameras}, are an attractive component of perception systems used in autonomous vehicles.
Hence, a number of works related to the topic discussed can be found in the literature.
The paper \cite{b51} presents the use of traditional convolutional neural networks for pedestrian detection.
The authors used their own method of representing event data in the form of frames called neighbourhood suppression time surface, which assumes that the intensity of each pixel on the time surface is only suppressed by its local neighbourhood.
The pedestrian detection system implemented in this way achieved an accuracy of 86\% AP (Average Precision). 



Another interesting approach in the context of object detection is the use of data fusion -- processing both event data and frames from a traditional camera. 
In the paper \cite{b52}, it was shown that this approach can significantly improve performance, especially in challenging lighting conditions and in the case of fast movement of objects or the camera.
A Spiking Neural Network (SNN) was used to enable event data to be processed directly.
The idea of data fusion was also applied in the paper \cite{b53}, this time using data from a traditional camera and a frequency representation of event data as input of the same network. 
Designed in this way, the detector achieved 81.6 \% AP for the DDD17 set \cite{ddd17}.



\subsection{Traffic sign detection}


In the present study, we used event data for single-class detection of traffic signs. 
This issue is very important in the context of advanced driver assistance systems and autonomous vehicles. 

The described detection problem is characterised by a relatively high complexity due to significant differences in the appearance of objects belonging to the same class. 
Traffic signs have different dimensions, shapes, and graphics on them. 
Moreover, a number of objects can be observed in road conditions that are similar to signs: car wheels, shop signs in urban conditions, etc.
This often is the cause of false positive detections.
The use of event cameras can potentially improve the quality of traffic sign detection in challenging lighting conditions (low light or high contrast), as well as in the case of fast movement of objects in relation to the camera. 

\section{Dataset}
\label{ch:dataset}

We use the Prophesee 1 Megapixel automotive detection dataset \cite{b6} in our research.
It consists of more than 15 hours of event data recorded in road conditions, divided into 60-second sequences. 
The collected event data is supplemented by annotations that take into account object classes characteristic of autonomous vehicles -- pedestrians, different vehicle classes, traffic signs and traffic lights.

\subsection{Data filtering}

\begin{figure}[!t]
    \centering
    \includegraphics[width=0.45\textwidth]{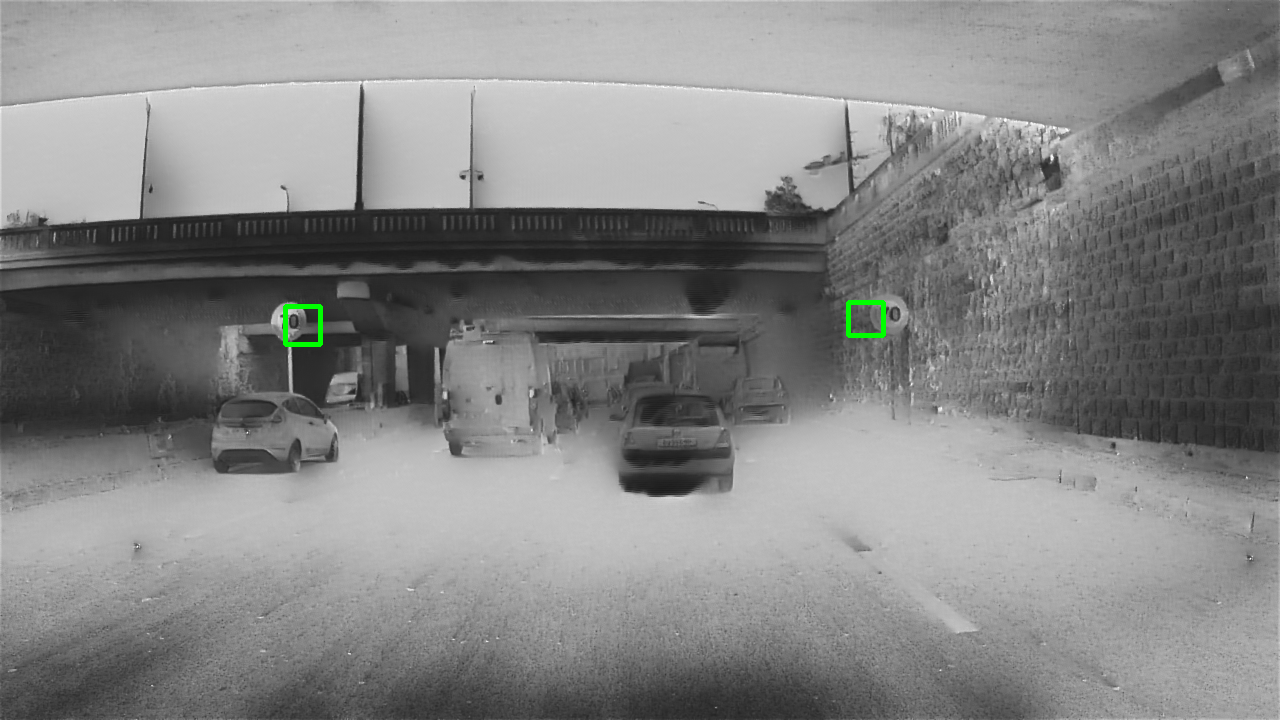}
    \caption{Example of a mislabelled traffic sign} 
    \label{fig:bad_miss}
\end{figure}

\begin{figure}[!t]
    \centering
    \includegraphics[width=0.45\textwidth]{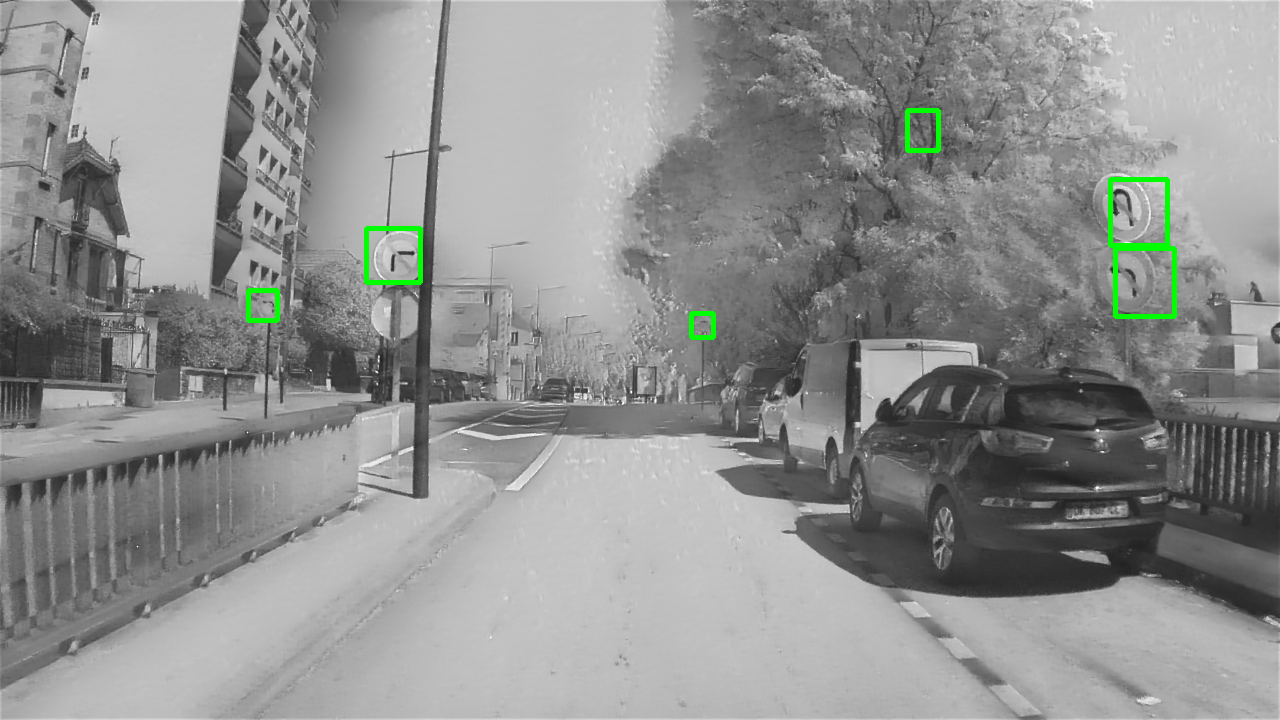}
    \caption{False positive example} 
    \label{fig:bad_false}
\end{figure}

The dataset used contains a lot of information that is not relevant to the traffic sign detection task. 
From the collected event data, we have filtered out exclusively the sequences in which there are traffic signs. 
For this purpose, we used the annotations provided by the authors of the dataset. 
Next, in order to visualise and verify the quality of the dataset, we performed a reconstruction of the event data into video frames using FireNet. 
After verifying different data accumulation times, we decided to use 10 ms time windows as a compromise between the increasing reconstruction time and the quality of the reconstructed frames. 
In this way, we obtained more than 1.4 million images with corresponding bounding boxes that marked the positions of the traffic signs. 
On the basis of manual verification of a portion of the set, we determined that the annotation provided by the authors of the dataset was of low quality -- not infrequently were the actual signs in a significantly different position than the corresponding bounding box (Figure \ref{fig:bad_miss}).
Additionally, the collection contained a large number of false positive detections in areas where no sign was visible (Figure \ref{fig:bad_false}).

\subsection{Automatic data verification}

Due to the large number of false positive detections and mislabellings of real objects within a given time window, we decided to perform an automatic verification of the data quality in the set used. 
For this purpose, we used the reconstructed video frames and a simple classifier based on a deep convolutional neural network. 
Our proposed solution consisted of 5 convolutional layers with ReLU (Rectified Linear Unit) activation functions, batch normalisation and MaxPooling, and 2 fully connected layers (dropout layers were additionally used). 
It was trained to classify objects as ``sign'' or ``no sign'', using the ADAM optimiser and the categorical crossentropy loss function. 
During the training process, we used 12663 images from the GTSRB (German Traffic Sign Recognition Benchmark) \cite{b7} dataset and 5379 images manually selected from the prepared frame reconstructions.
After 25 epochs of training the network for an input size of 64x64 pixels, we obtained a classifier with a performance of 95 \% on the test set.

\subsection{Final dataset}

We used the tool described in the previous section to select the images reconstructed from the dataset that contain correctly annotated traffic signs. 
In this way, we obtained 27629 reconstructed images, which were identified by the classifier as correctly labelled. 
The proposed solution significantly reduced the number of false positive cases.

Based on the selected reconstructed images from the corresponding 10 ms event time windows, we prepared additional data representations described in Section \ref{ch:representations}: 3 independent single-channel event representations and a three-channel fusion of all of them. 
Each representation included correctly labelled traffic signs. 
We then used the data to implement the traffic sign detection systems based on event data. 

It is worth noting that the prepared dataset still contains problematic cases.
The used tool does not allow one to eliminate false negative cases and this will be addressed in our future research. 
However, the set prepared in this way is characterised by significantly higher quality than the original one.




\section{The proposed detection system}
\label{ch:train}
The deep convolutional neural network YOLOv4 was chosen as the detector for this task \cite{b8}, because of its high accuracy and low inference time.
Both proprieties are very important in the automotive context.
It is a very commonly used solution, so it provides a good platform for studies comparing detection performance for different types of input data.
 
 
For the YOLOv4 model, we applied the $CIoU$ loss function. 
We used DropBlocks for regularisation, and $MOSAIC$ augmentation. 
The YOLOv4 network training process was performed using $4$ {GPU NVIDIA V100} cards for an input size of $640x640x3$ and a batch size of $64$. 
In order to reduce training time, we used a transfer-learning technique (using weights adapted to detect objects for the \emph{MS COCO} dataset \cite{ref_mscoco}) and high-performance hardware resources provided by the PLGrid infrastructure and the CYFRONET Academic Computer Centre computing cluster.


We have prepared 5 independent DCNN models -- for each of them, analogous parameters, input size, and architecture were used. 
The learning process was carried out for equal datasets (a division into a test set and a learning set in the ratio of 0.25:0.75 was established). 
For each network, we used representations corresponding to the same events from the dataset. 

The research started with the process of training networks using one of the three considered representations as input: event frame, event frequency, and the exponentially decaying time surface. 
For each representation we achieved a rather satisfactory accuracy in the range of 86.9-88.9\% mAP@0.5 (as shown in Table \ref{tab:result}).

Then we prepared a detector using the fusion of the three considered representations as consecutive input channels. 
The motivation behind the proposed representation fusion was to use as much information as possible from the source event data. 
During the learning process, after 26500 iterations we achieved a detector characterised by an accuracy of 89.9\% mAP0.5 (mean average precision with intersection over union threshold 0.5) 
To compare the results obtained, we also performed a training process for video frames reconstructed with FireNet as CNN input. 
After 13600 iterations, we achieved a performance value of 72.67\% mAP@0.5. 
A comparison of all the mentioned approaches is summarised in Table \ref{tab:result}.

We also tested the prepared detectors in terms of throughput. For the Jetson AGX Xavier eGPU platfrom we achieved an average of 7.7 $fps$ (frames per second) for a few second video sequences. The throughput of the solution is crucial due to the application of detectors for automotive purposes, and increasing it is the object of further planned research. It can be improved by using TensorRT \cite{tensorrt} optimisation for the eGPU or quantisation and hardware acceleration for the FPGA platform.

\begin{table}[!t]
    \centering
    \caption{Achieved detection quality measures for each selected YOLOv4 input. Our approach to apply a fusion of selected representations made it possible to obtain a detector with the highest accuracy. On the other hand, the lowest accuracy was achieved for the frames reconstructed using FireNet.}
    \label{tab:result}
    \begin{tabular}{|l|c|c|c|c|}
        \hline
        \emph{Input} & \emph{Precision} & \emph{Recall} & \emph{F1-score} & \emph{mAP@0.5}  \\
        \hline
        \emph{Decaying time space} & 0.94 & 0.81 & 0.87 & 86.9\% \\
        \hline
        \emph{Event frequency} & 0.94 & 0.84 & 0.89 & 88.67\% \\
        \hline
        \emph{Event frame} & 0.93 & 0.82 & 0.87 & 86.93\% \\
        \hline
        \emph{\textbf{Fusion}} & \textbf{0.94} & \textbf{0.86} & \textbf{0.90} & \textbf{89.9\%} \\
        \hline
        \emph{Reconstruction} & 0.97 & 0.72 & 0.83 & 72.67\% \\
        \hline
    \end{tabular}
\end{table}

The study shows that the event data representation method used in the training process does affect the quality of the detection system. 
The fusion of representations that use different features of event data is characterised by higher amount of information and, consequently, makes it possible to achieve higher object detection accuracy.
The frames reconstructed using FireNet's deep neural network have the characteristics of images recorded by traditional cameras and provide an accessible form of representation of the scene captured by the event camera. 
However, blurring and the limited level of detail of the frames result in only moderate performance of the object detection task.
\begin{figure}[!t]
    \centering
    \includegraphics[width=0.45\textwidth]{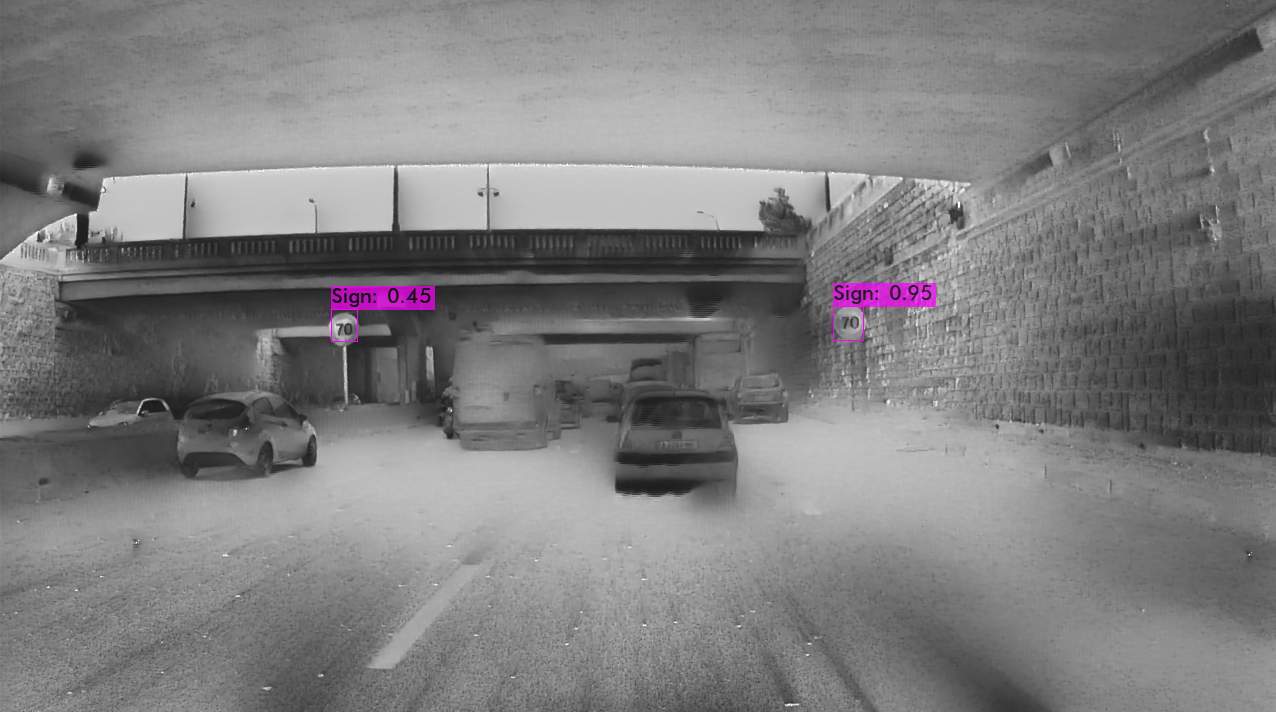}
    \caption{Example of correct detection on frames reconstructed with FireNet} 
    \label{fig:reconstructed_detect}
\end{figure}

\begin{figure}[!t]
    \centering
    \includegraphics[width=0.45\textwidth]{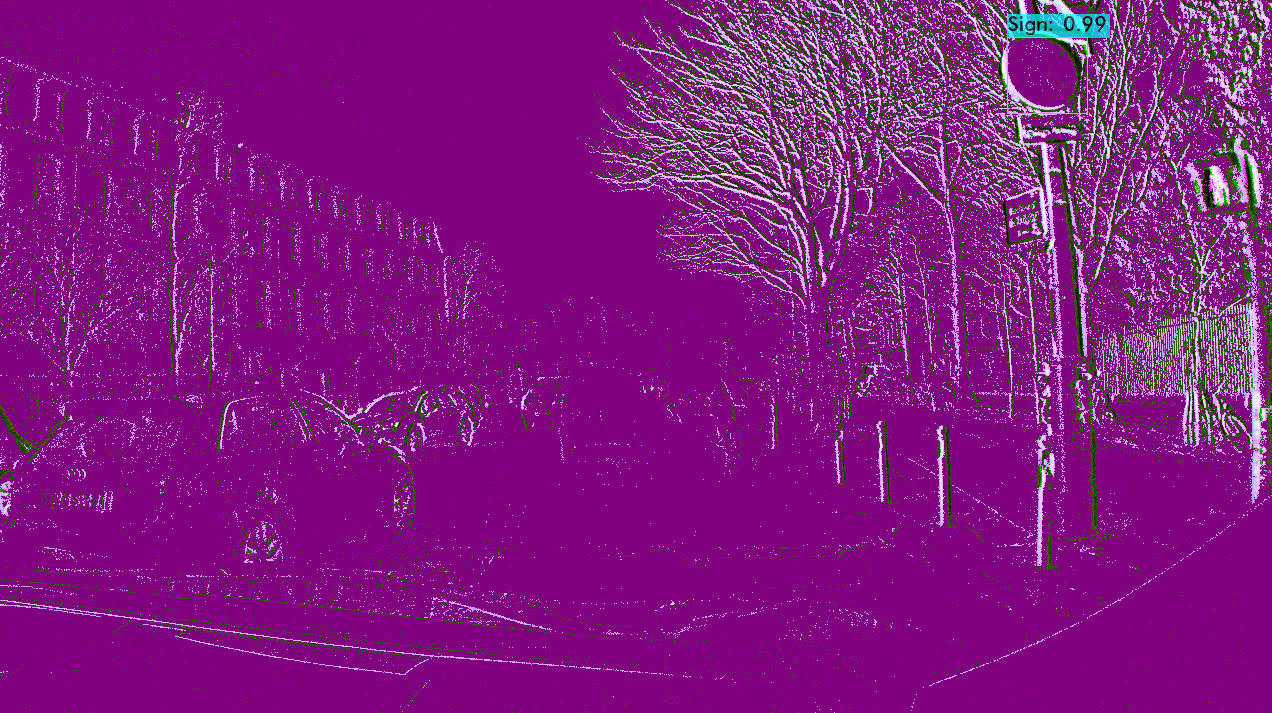}
    \caption{Example of correct detection on the fused representations} 
    \label{fifusion_detect}
\end{figure}

\section{Summary}
\label{ch:summary}

In this paper, we have evaluated the possibility of using event data for traffic sign detection using a YOLOv4 network. 
The specific event data format (sparse point clouds in space-time) and the desire to use a typical deep neural network necessitated the use of i.e. event data representation or reconstruction.
Experiments have shown that the use of single-channel representations, such as event frame, event frequency, and exponentially decaying time surface, does enable the task of traffic sign detection to be performed with a rather satisfactory accuracy in the range of 86.9-88.9\%@mAP. The use of fusion, i.e. the combination of individual representations into a three-channel image, made it possible to obtain a detector characterised by a higher accuracy of 89.9\% mAP@0.5.
It is worth noting that individual representations take into account other features of the event data, which made it possible to obtain a synergy effect.
The last solution considered was the use of the FireNet video frame reconstruction network. 
Based on the learning data prepared in this way, we obtained a detector characterised by an accuracy of 72.67\% mAP0.5.
The significantly lower performance can be justified by the atypical nature of the data so obtained, namely, the frames reconstructed by the network have the characteristics of images captured by classical cameras, but are accompanied by significant blurring and relatively little detail.

Using event data and classical deep convolutional networks, which in the context of computer vision are considered state-of-the-art for detection and many other tasks, is possible, effective and efficient -- data aggregation and representation are based on operations with low computational complexity, and networks for frame reconstruction have surprisingly small sizes.
The prepared detection systems are characterised by a rather high accuracy despite the low quality of the used dataset and high complexity of the problem considered.
However, we should point out that implementing traffic sign recognition with the same approach could be very challenging, as here fine details presented in the sign are important.  
The research conducted underlines the potential of event cameras for automotive applications, as a stand-alone sensor or as a support for traditional cameras.



As part of further work, first of all we plan to refine the dataset analysis methodology and thus eliminate all problematic cases.
In parallel, we also want to acquire our own sequences (simultaneously from an event and a traditional camera), also in very demanding situations such as driving at night, entering and leaving a tunnel or during sunrise and sunset.
Another direction of research is the fusion of video and event data, which should improve accuracy and enable the implementation of a complete traffic sign recognition system.
We are also planning a hardware implementation of the solution for SoC FPGA (System on Chip Field Programmable Gate Arrays) platforms, which will enable real-time operation of the vision system.



\section*{Acknowledgment}

The work presented in this paper was supported by the AGH University of Science and Technology project no. 16.16.120.773 and Silesian University of Technology. Access to high-performance hardware was possible due to the use of the CYFRONET Academic Computer Centre computing cluster.

\end{document}